\documentclass[a4paper,10pt,twocolumn]{article}
\usepackage{xcolor}
\usepackage[utf8]{inputenc}
\usepackage{graphicx}
\usepackage{amsmath}
\usepackage{amssymb}
\usepackage{geometry}
\usepackage{fancyhdr}
\usepackage{titlesec}
\usepackage{helvet}
\usepackage{mathptmx}
\usepackage{hyperref}
\usepackage[backend=bibtex, style=ieee]{biblatex}
\usepackage{hyperref}
\usepackage[nolist]{acronym}
\usepackage[capitalize]{cleveref}
\begin{acronym}[H.264/SVC]
    \acro{FL}{Federated Learning}
    \acro{CNN}{Convolutional Neural Network}
    \acro{GRU}{Gated Recurrent Unit}
    \acro{LSTM}{Long-Short Term Memory Network}
\end{acronym}

\fancypagestyle{firstpageheader}{
    \fancyhf{}
    \fancyhead[L]{\includegraphics[height=30pt]{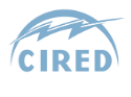}} 
    \fancyhead[C]{\begin{tabular}{c}
        \textbf{CIRED Workshop on Resilience of Electric Distribution Systems} \\
        Chicago, November 7-8, 2024
    \end{tabular}}
    \fancyhead[R]{\textbf{Paper n° 199}}
    
}

\titleformat{\section}{\bfseries\uppercase}{\thesection.}{1em}{}
\titleformat{\subsection}{\bfseries}{\thesubsection.}{1em}{}
\titleformat{\subsubsection}{\itshape}{\thesubsubsection.}{1em}{}

\title{\textbf{Federated Learning Forecasting  for Strengthening Grid Reliability and Enabling Markets for Resilience}}
\author{
  Lucas PEREIRA$^1$
  \and
  Vineet NAIR$^2$
  \and
  Bruno DIAS$^3$
  \and
  Hugo MORAIS$^4$
  \and
  Anuradha ANNASWAMY$^2$
}
\date{$^1$ITI/LARSyS, IST, Portugal | $^2$MIT, USA | $^3$UFJF, Brazil  | $^4$INESC-ID, IST, Portugal}

\begin{acronym}
    \acro{DER}{Distributed Energy Resource}
    \acro{FL}{Federated Learning}
    \acro{LV}{Low Voltage}
\end{acronym}

\begin{document}

\maketitle
\thispagestyle{firstpageheader}

\begin{abstract}
We propose a comprehensive approach to increase the reliability and resilience of future power grids rich in distributed energy resources. Our distributed scheme combines federated learning-based attack detection with a local electricity market-based attack mitigation method. We validate the scheme by applying it to a real-world distribution grid rich in solar PV. Simulation results demonstrate that the approach is feasible and can successfully mitigate the grid impacts of cyber-physical attacks.
\end{abstract}

\section{Introduction and background}
The variability and intermittency associated with wind and solar introduce more challenges to balance supply and demand and ensure reliable grid operation. In addition, increasing penetration of batteries, electric vehicles, and flexibilities introduces more complexity and uncertainty in net-load forecast \cite{denholm_overgeneration_2015}. Furthermore, a grid rich in \acp{DER} is more vulnerable to cyber-physical attacks. Various types of attacks on DER-rich power grids have been described in the literature \cite{soltan2018blackiot,shekari2022madiot}.

Uncertainties can impact several tasks needed in power systems, such as security assessment, operational planning, wholesale electricity markets, hosting capacity, and resiliency strategies. This can result in violations of technical operating constraints, non-compliance with market rules, imposing unnecessary limits on hosting capacity, or failure to identify attacks \cite{hassanzadeh2023improving}. In this sense, accurate forecasting with uncertainty quantification is crucial to successfully detecting and mitigating these issues.

The Accurate federated Learning with uncertainty quantification for \ac{DER} forecasting Applied to sMart Grids planning and Operation (ALAMO \cite{pereira_accurate_2024}) project focuses on developing technologies to manage power grids with high penetration of \acp{DER} while ensuring stakeholder privacy. It aims to create accurate forecasting algorithms based on \ac{FL} and address challenges in quantifying epistemic and aleatoric uncertainties. The project will explore recent \ac{FL} variations of client selection and model aggregation to enhance forecasting accuracy, which is currently inferior to traditional centralized models. Additionally, it aims to develop and benchmark uncertainty quantification methods, like Quantile-Regression and deep-model ensembles, to achieve sharp and well-calibrated uncertainty estimates. 

With this paper, we aim to leverage the ALAMO project framework to show that by combining accurate probabilistic forecasts and uncertainty estimates with a market mechanism, we can enhance grid reliability during nominal operation and provide resilience to various cyber-physical attacks.

This paper details the proposed framework to include FL in the forecasting tasks and market mechanisms. A use case with a five-feeder real-world \ac{LV} distribution network will be presented to demonstrate the feasibility of the proposed solution. The paper is organized as follows. We briefly overview \ac{FL} in \cref{sec:fl_intro}. We then describe the methodology in \cref{sec:methods}, including the FL-based attack detection and market-based attack mitigation. \cref{sec:sim} presents the use case, datasets, FL forecasting implementation, and some simulation results. Finally, we summarize conclusions in \cref{sec:sim} along with some ideas for future work.

\subsection{Federated Learning \label{sec:fl_intro}}
\ac{FL} is a decentralized approach that enables collaborative training of machine learning models across distributed environments where data is stored locally on different organizational devices or systems. This method enhances privacy by ensuring that sensitive data remains on local devices, addressing privacy concerns inherent in centralized models. Additionally, FL improves model training efficiency and scalability by reducing storage costs and communication burden, as it transmits data wirelessly without needing a physical connection \cite{Tan2023,QI2024}. \ac{FL} is particularly beneficial in the energy systems domain, as it allows information from various network points to be used during training, resulting in more accurate predictions.

For example, for household demand forecasting, \cite{Taik2020} and \cite{Savi2021} both utilize FL with \ac{LSTM} networks and the FedAvg aggregation method. These approaches have demonstrated efficient network resource use, reduced training time, and enhanced privacy preservation. Additionally, 
\cite{GeZhang2022} explores FL with \ac{CNN} and LSTM models, comparing various aggregation strategies, with FedAdagrad showing resilience against false data attacks. In the realm of PV production forecasting, \cite{HOSSEINI2023116900} presents an FL model using a multi-layer perceptron, achieving high accuracy while ensuring data privacy. \cite{MUHAMMAD2024} integrates CNN and \ac{GRU} networks with FL and the Orchard Algorithm, resulting in superior prediction performance. \cite{MORADZADEH2023} introduces a federated deep learning model for PV power generation, demonstrating robustness against cyber attacks and generalizability across regions.

\section{Methods}\label{sec:methods}

\subsection{Using FL Forecasts to Identify Cyber Attacks\label{sec:methods_fl}}
Cyber-physical attacks on power grids can broadly be classified into deception, disclosure, and disruption attacks that compromise integrity, confidentiality, and resource availability, respectively \cite{dibaji2019systems}. Here, we focus specifically on denial of service (DoS) disruption attacks that directly disconnect resources from the network. 

In this work, the FL paradigm is used to obtain day-ahead forecasts of household demand and PV production in each individual prosumer in the grid. The forecasts are then used to identify attacks using a threshold-based method that leverages the individual forecast errors and the difference between the power import at the feeder (assuming that aggregated measurements are available) and the aggregation of the forecasts of the connected prosumers. On the one hand, by studying the forecast errors of demand and PV production in each prosumer, it is possible to identify any drastic deviations that occur only in particular nodes. On the other hand, it is possible to identify any unexpected deviations by comparing the power import at the feeder with the aggregated predictions on each feeder. This allows us to detect anomalies and flag whether an attack has occurred. 

A few prior works have applied \ac{FL} for anomaly detection \cite{jithish2023distributed} and the detection of other cyber attacks like false data injection \cite{li2022detection}. However, to our knowledge, ours is the first work to combine the distributed FL approach with a distributed market structure for attack detection and mitigation and apply this to a truly DER-rich grid.

\subsection{Local Electricity Markets}\label{sec:methods-mkts}

\begin{figure}[htb]
\centering
\includegraphics[width=0.8\columnwidth]{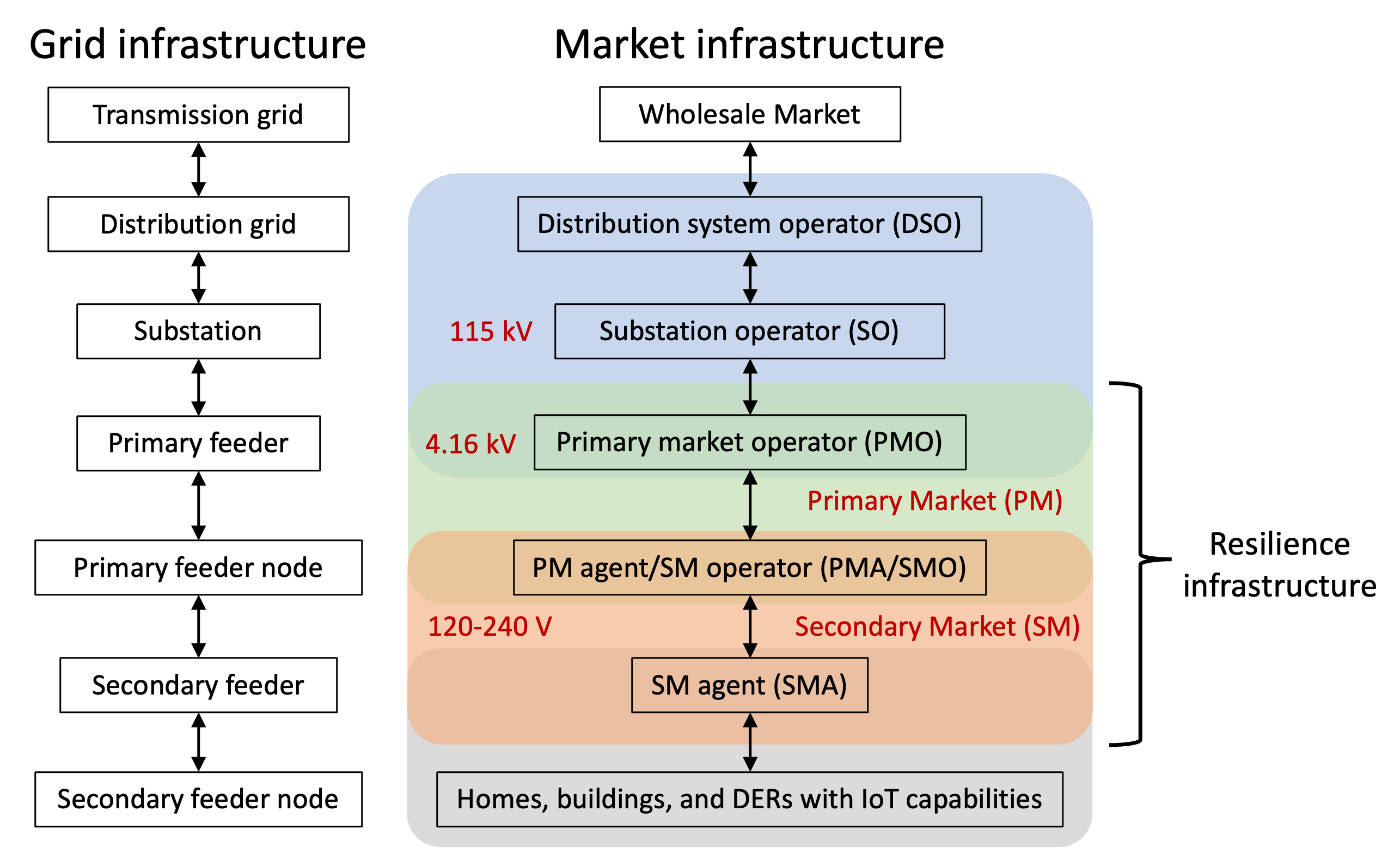}
\caption{Overview of hierarchical LEM, with the PM and SM layers utilized for resilience.
\label{fig:mkt_agents}}
\end{figure}

\begin{figure}[htb]
\centering
\includegraphics[width=0.95\columnwidth]{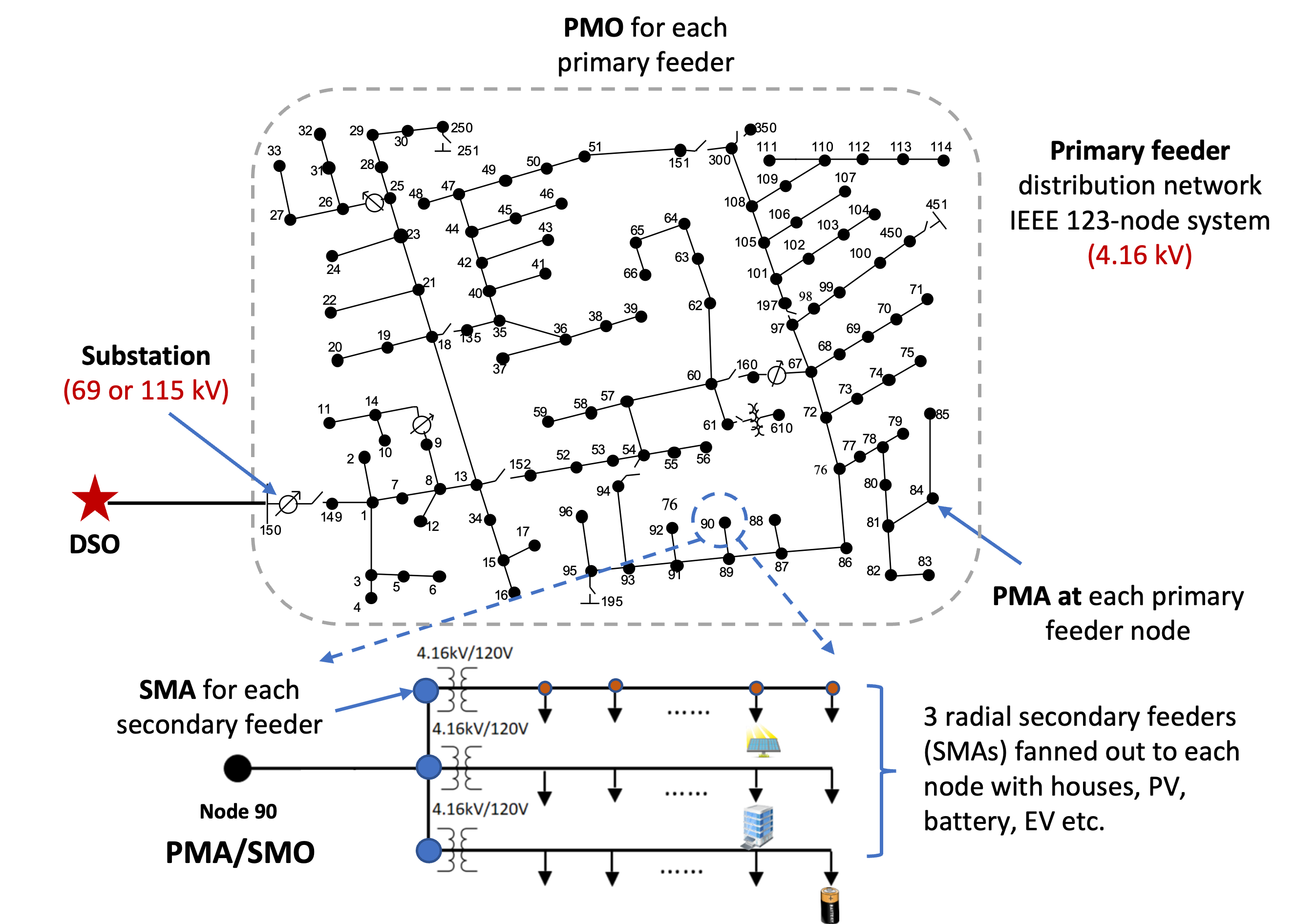}
\caption{LEM co-located with distribution grid. This shows a primary and secondary feeder distribution network based on the modified IEEE-123 node test case.\label{fig:network_schematic}}
\end{figure}

We leverage a hierarchical local electricity market (LEM) structure developed in our prior work \cite{nair2022hierarchical}. The LEM is described in \cref{fig:mkt_agents} and \cref{fig:network_schematic}. In previous papers, we have applied this to provide grid services like voltage control \cite{nair2023local} and enhance grid resilience to cyber-physical attacks \cite{nair2024enhancing}.
Here, we combine this market with FL to detect and mitigate attacks on different assets in the grid. We focus only on the primary market in this work. Each node in the feeder represents a PM agent (PMA) and has a house with rooftop solar PV and loads connected to it. The PM for the entire feeder is overseen by a PM operator(PMO). After receiving flexibility bids from each PMAs, the PM is cleared by solving an alternating current optimal power flow (ACOPF) problem. Since the distribution network is 3-phase unbalanced and radial, we apply a current injection (CI)-based power flow model that captures all the grid physics \cite{ferro2020distributed}. The ACOPF minimizes line losses, generation costs, and disutility due to load flexibility. The problem is solved using a distributed optimization algorithm known as proximal atomic coordination \cite{romvary2021proximal}.

\subsection{Mitigation Approach Using Load Flexibility}
After detecting an attack, the PMO raises a flag and commences mitigation efforts. It does so by comparing the actual meter reading with the attack ($\overline{\textbf{P}}_{PCC}$) vs the forecasted value without the attack ($\textbf{P}_{PCC}$) for the 3-phase net power injection at the substation or PCC (connected to the main grid via a tie line). It uses this information to update the cost coefficients for each of the terms in the objective function. The update rule is specified in \cref{eq:update_ci_model}. Note that we use hour-ahead forecasts for the power injections at a 1-minute resolution for the market operation.
\begin{gather}
\Delta=\mathbf{P}_{PCC}-\overline{\mathbf{P}}_{PCC} \\
 Z_i\left(\delta_i\right)=1+\frac{RS_i \Delta^{\top} \delta_i}{\mu \sum_i RS_i} \Longrightarrow \gamma_{i \delta}=\frac{1}{Z_i\left(\delta_i\right)} \\
\overline{\boldsymbol{\alpha}}_i=\gamma_{i \alpha} \boldsymbol{\alpha}_i, \quad \overline{\boldsymbol{\beta}}_i=\gamma_{i \beta} \boldsymbol{\beta}_i, \quad \overline{\boldsymbol{\xi}}=\left(\frac{\sum_i \gamma_{i \alpha}+\gamma_{i \beta}}{2n}\right)^{-1} \boldsymbol{\xi} \label{eq:update_ci_model}
\end{gather}
$\boldsymbol{\alpha}_i, \boldsymbol{\beta}_i$ are $3\times 1$ vectors representing cost and disutility coefficient for each phase at SMO node $i$, and $\boldsymbol{\xi}$ is a 3-phase hyperparameter that penalizes line losses in the objective function. A distributed generation attack that increases net load would result in $|\overline{\textbf{P}}_{PCC}| > |\textbf{P}_{PCC}|$ and $\gamma_{i\alpha}, \gamma_{i\beta} < 1$ and $\overline{\boldsymbol{\xi}} < \boldsymbol{\xi}$. By artificially lowering local generation costs and load disutility parameters, along with penalizing line losses more heavily, this update results in a dispatch that favors more local DER generation and load flexibility instead of relying on transmission imports. A key difference here is that the PRM also takes into account the RS of each SMO during the redispatch so that it relies more heavily on resilient SMOs for attack mitigation. Note that the PMO also accounts for the resilience scores (RS) of the PMAs while updating the coefficients, so as to rely more heavily on more resilient and reliable PMAs to provide flexibility during attack mitigation \cite{nair2024resilience}. The PMO updates the new coefficients $\alpha_i^\prime, \beta_i^\prime$ and $\xi^\prime$, and sends these to all the PMAs. The PM is then re-dispatched by solving the ACOPF again to assign new set points to the PMAs.

\section{Simulation Case Study}\label{sec:sim}

\subsection{Low Voltage Distribution Grid}

The simulation was conducted considering a real-world Low Voltage (LV) distribution network in Madeira Island. The secondary substation that feeds this network has a transformer with an apparent power of 250 kVA, connected in delta-wye, which transforms voltage from the transmission grid (6600 V) to the distribution grid (400 V) \cite{faustine_fpseq2q_2022}. This radial LV network has 88 nodes connected through 87 lines. The system was originally modeled using DIgSILENT PowerFactory \cite{ponnaganti_battery_2020}, from which the admittance matrix was later calculated.

\subsection{Consumption and PV Production Data}
Since the actual measurements are unavailable for each node, we relied on consumption and production data from 12 prosumers from Madeira Island. This data was collected during the Horizon 2020 SMILE project and is available at 1 sample per minute (1/60 Hz) \cite{pereira_economic_2020}. Moreover, each of the 88 nodes in the LV grid is assumed to correspond to a single-phase prosumer. The consumption and PV production profiles were randomly assigned to one phase in each node. We synthetically generated PMA flexibility bids at each node by randomly assigning downward flexibilities between 20-40\% to each.

\subsection{Federated Forecasting}
For the federated forecasting, we relied on the FLOWER framework\footnote{\url{https://flower.ai/}} and the FPSeq2Quant probabilistic forecasting algorithm \cite{faustine_fpseq2q_2022}. The models were trained with one year of data for each of the 12 prosumers using a time-series split cross-validation with an expanding training window. The model averaging was performed using the standard Federated Averaging \cite{mcmahan2017}. More precisely, four federated forecasting models were trained: 1) day-ahead demand, 2) day-ahead PV production, 3) hour-ahead demand, and 4) hour-ahead PV production. The day-ahead forecasts were developed considering 15-minute aggregated values, whereas the hour-ahead forecasts consider samples every 1 minute. 

\subsection{Experiments and Results}
We simulated the LEM over a 24-hour period. We considered that the attack occurs in the middle of the day (12:30 pm) during the period of peak PV output to maximize attack impact. In our case study, we considered that all the rooftop PV generation units have been attacked and shut down. These can be taken offline by hacking the smart inverters that connect these assets to the grid. For simplicity, we only show results for an instant period of time during the attack period. 

\begin{figure}[htb]
\centering
\includegraphics[width=0.75\columnwidth]{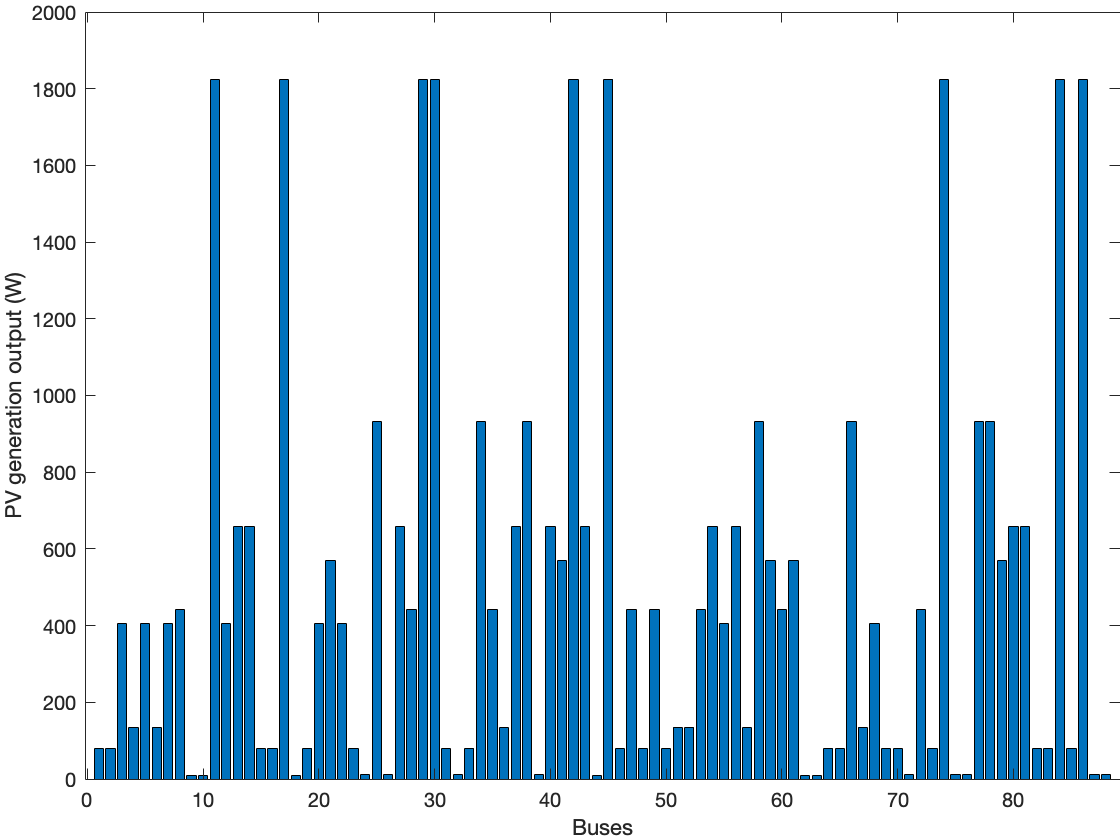}
\caption{PV generation before the attack.\label{fig:gen}}
\end{figure}

\begin{figure}[htb]
\centering
\includegraphics[width=0.7\columnwidth]{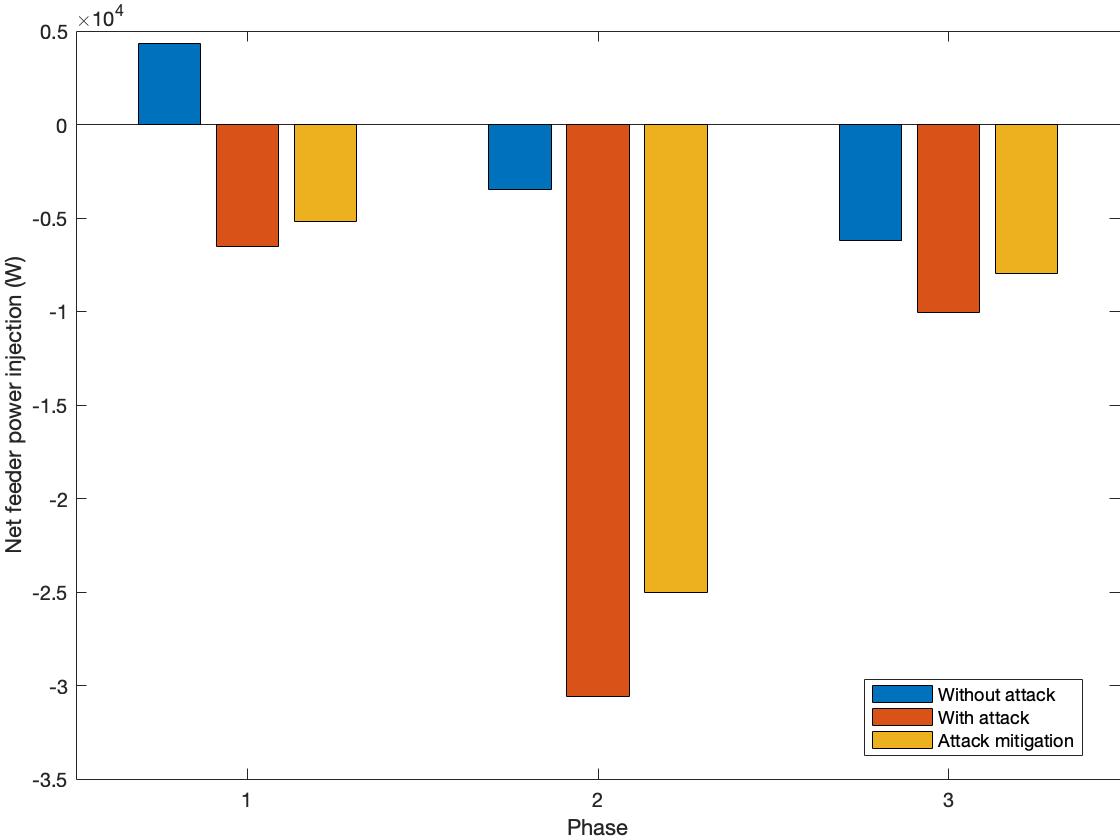}
\caption{Net total feeder power injection (3-phase). \label{fig:feeder}}
\end{figure}

\begin{figure}[htb]
\centering
\includegraphics[width=0.75\columnwidth]{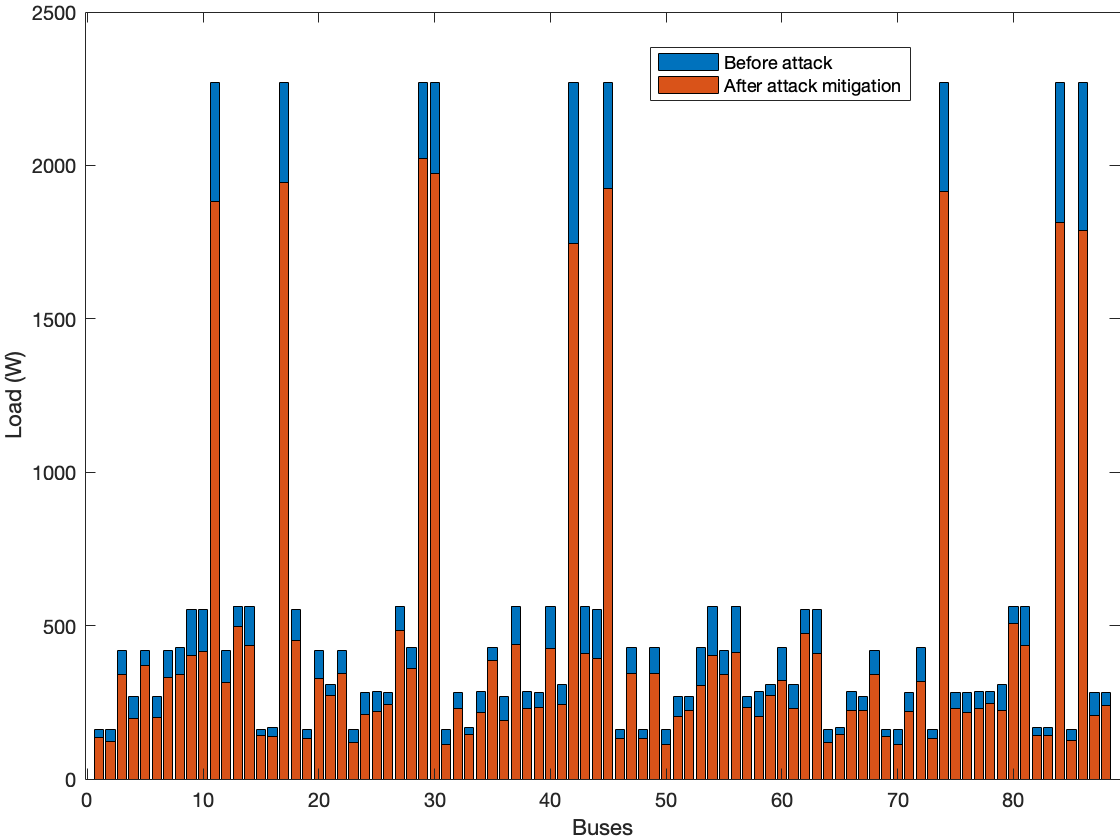}
\caption{Distribution of load curtailment across nodes.\label{fig:load_curtail}}
\end{figure}

\cref{fig:gen} shows the PV generation over all 88 nodes before the attack. After the attack, this reduces to zero. This is a total loss of 100\% of generation. This increases the power import at the feeder from the main grid. As seen in \cref{fig:feeder}, the feeder exports power to the grid on phase A before the attack. However, after the attack, all 3 phases draw power from the grid, and the total net-load increases from 5.4 kW to 47.2 kW. The PMO then applies the update-based mitigation by leveraging load flexibility. The distribution of load curtailment across all the PMA nodes is shown in \cref{fig:load_curtail}. Thus, we reduced the power import to about 29 kW, closer to the level without the attack. However, due to the relatively large scale of the attack and limiting power flow network constraints, we cannot fully resolve it even after utilizing most of the load flexibility available. Thus, the power import is still higher than before but about 40\% lower than the case without mitigation. This ensures we minimize the distribution grid attack’s impact on the larger transmission grid. This can help prevent more widespread impacts like frequency instability and cascading failures.

\section{Conclusions and Future Work}\label{sec:conc}

In this paper, we proposed a method integrating federated learning with local energy markets to detect and mitigate cyber-physical attacks to DER-rich LV networks.

Still, several challenges must be addressed to advance this work, especially concerning training such \ac{FL} models. 
In this sense, efficient communication is crucial because federated networks with many local devices can be slower than local computing. Additionally, the statistical heterogeneity of data from different devices increases delays in data processing, complicating modeling, analysis, and assessment \cite{article-zhu-2021,article-qinbin-2022}. We aim to tackle these issues and improve our FL models and workflows in future work.

We will also explore more sophisticated attack detection approaches beyond our simple threshold-based method. Finally, we will make our market operation more realistic by also incorporating forecast uncertainty (for both PV and load) into the problem through approaches like robust optimization, stochastic programming and/or chance constraints.

\section*{Acknowledgement}
FCT supports this work through projects 10.54499/2022.15771.MIT; 10.54499/LA/P/0083/2020, 10.54499/UIDP/50009/2020, 10.54499/UIDB/50009/2020, and grant 10.54499/CEECIND/01179/2017/CP1461/CT0020 (LP).

\vspace{10px}
\noindent CIRED Workshop on Resilience of Electric Distribution Systems, Chicago, November 7-8, 2024, Paper n° 118
\printbibliography
\end{document}